\documentclass{article}

\usepackage{PRIMEarxiv}

\usepackage[utf8]{inputenc} 
\usepackage[T1]{fontenc}    
\usepackage{hyperref}       
\usepackage{url}            
\usepackage{booktabs}       
\usepackage{amsfonts}       
\usepackage{nicefrac}       
\usepackage{microtype}      
\usepackage{lipsum}
\usepackage{siunitx}
\usepackage{adjustbox}
\usepackage{amsmath}        
\usepackage{fancyhdr}       
\usepackage{graphicx}       
\usepackage{nicefrac}       
\usepackage{afterpage}
\usepackage{appendix}
\graphicspath{{media/}}     
\usepackage{multirow}
\usepackage{wrapfig,lipsum,booktabs}
\usepackage{caption}
\usepackage{subcaption}
\usepackage[table,xcdraw]{xcolor}
\usepackage[justification=centering]{caption}

\usepackage[table,xcdraw]{xcolor}
\usepackage[normalem]{ulem}
\useunder{\uline}{\ul}{}

\title{SCARED-C: Corrected Camera Poses for Endoscopic Depth Estimation} 

\author{
    John J. Han \\
        Vanderbilt University \\
        Nashville, TN, USA \\
    \And 
    Adam Schmidt \\
        Intuitive Surgical, Inc. \\
        Sunnyvale, CA, USA \\
    \And 
    Max Allan \\
        Intuitive Surgical, Inc. \\
        Sunnyvale, CA, USA \\
    \And 
    Jie Ying Wu \\
        Vanderbilt University \\
        Nashville, TN, USA \\ 
    \And 
    Omid Mohareri \\ 
        Intuitive Surgical, Inc. \\
        Sunnyvale, CA, USA
}

\begin{document}
\maketitle

\begin{abstract}
The SCARED dataset~\cite{scared} is a widely used benchmark for endoscopic depth estimation, offering ground-truth 3D reconstructions captured with a structured light sensor. However, the depth maps for non-keyframe images rely on robot kinematics that introduce substantial pose errors, limiting the reliably labeled portion of the dataset to 35 keyframes. We present SCARED-C, a corrected version of the SCARED dataset that expands the number of reliable RGB-D pairs from 35 to 17,135. Our pipeline applies COLMAP~\cite{colmap}, a Structure-from-Motion system, to re-estimate camera poses for all frames, followed by a scale recovery step that aligns the resulting reconstructions to metric space using the ground-truth keyframe depth maps. We validate the corrected poses through (1) stereo disparity evaluation and (2) monocular depth estimation experiments. The corrected dataset and code are publicly released to the community.\footnote{\url{https://huggingface.co/datasets/juseonghan/SCARED-C}}
\end{abstract}

\section{Introduction}

Depth estimation is a fundamental task in surgical computer vision, with applications ranging from augmented reality overlays to autonomous surgical assistance~\cite{han2026depth,acar2025monocular}. However, collecting ground-truth depth in a clinical endoscopic setting remains difficult due to the physical constraints of the operating environment. Existing labeled datasets for surgical depth estimation therefore rely on either simulated data~\cite{endoslam, realsyncol}, phantom scenes registered to 3D scans~\cite{c3vd, c3vdv2}, or, more recently, synthetic data generated through generative models~\cite{martyniak2025simuscope} and neural rendering techniques~\cite{han2025endopbr}.

Among real-tissue datasets, the SCARED dataset~\cite{scared} occupies a unique position. Introduced as part of a sub-challenge of EndoVis at the MICCAI 2019 conference, it provides ground-truth depth maps for ex-vivo porcine abdominal scenes captured using a structured light sensor mounted on a da Vinci endoscope. Each \textit{keyframe} consists of an RGB image paired with a depth map derived from the structured light reconstruction. Because the sensor can only capture depth from a static viewpoint, the dataset was extended by moving the endoscope arm and projecting the depth map of keyframe into the neighboring video frames using the robot's forward kinematics. In theory, this yields RGB-D supervision across entire video sequences.

\begin{figure}
    \centering
    \includegraphics[width=0.8\linewidth]{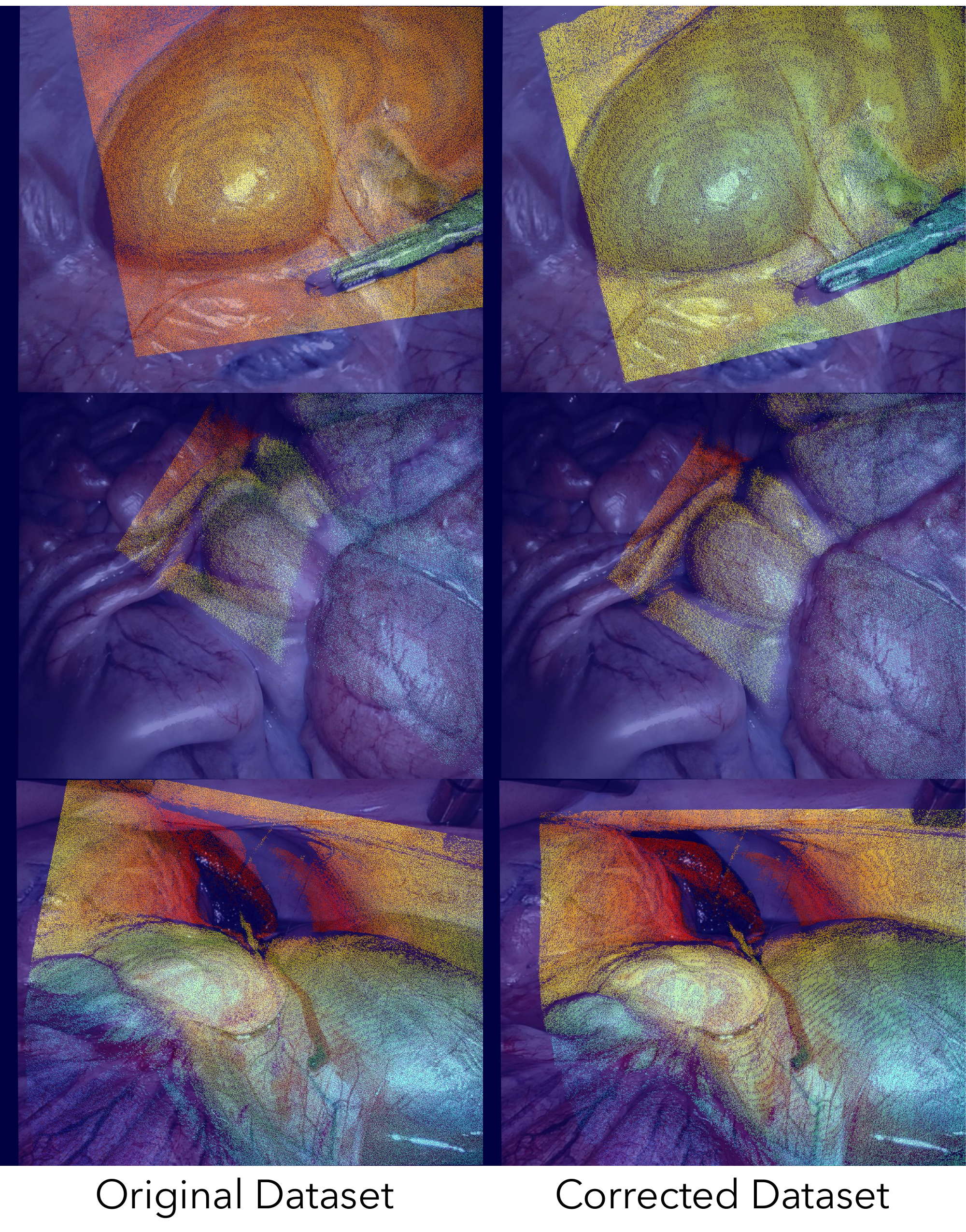}
    \caption{Samples from the original SCARED dataset (left) and the corrected SCARED-C dataset (right). The corrected depth maps exhibit substantially better alignment with the corresponding RGB images.}
    \label{fig:qual}
\end{figure}

In practice, however, the non-keyframe depth maps are unreliable. The da Vinci system is cable-driven to maintain a compact form factor, and this design introduces non-negligible kinematics errors. As noted by the original challenge authors~\cite{scared}, the resulting depth maps exhibit severe misalignment with their corresponding RGB images (see Fig.~\ref{fig:qual}), rendering most of the extended dataset unsuitable for training or evaluation. Consequently, prior work on the SCARED dataset has been limited to the 35 keyframes for which structured light ground truth exists.

We address this limitation by correcting the camera poses for all frames in the SCARED dataset. Rather than relying on robot kinematics, we use COLMAP~\cite{colmap}, an off-the-shelf Structure-from-Motion (SfM) pipeline, to estimate camera poses directly from the image data. Because monocular SfM recovers geometry only up to an unknown scale, we introduce a simple scale recovery algorithm that uses the ground-truth keyframe depth maps to transform the reconstruction into metric space. By reprojecting keyframe depth maps through the corrected poses, we obtain 17,135 reliable RGB-D pairs, resulting in roughly $490\times$ expansion over the original 35 keyframes.

Our contributions are as follows:
\begin{enumerate}
    \item We correct the non-keyframe camera poses in the SCARED dataset using Structure-from-Motion and a scale recovery algorithm, expanding the reliably labeled data from 35 to 17,135 frames.
    \item We validate the corrected dataset through two experiments: evaluation with an off-the-shelf stereo model (FoundationStereo~\cite{wen2025foundationstereo}) and a monocular depth estimation training comparison.
    \item We publicly release the corrected dataset on HuggingFace with a suggested train-validation split and release our scale-recovery code.
\end{enumerate}

\section{Method}

\subsection{Camera Pose Estimation via COLMAP}

For each video sequence in the SCARED dataset, we run COLMAP on the left camera frames at their original resolution. Because the da Vinci system produces natively undistorted images, we do not apply any additional undistortion. Since the optical center of the left camera does not coincide with the image center, we initialize COLMAP with the provided camera intrinsics and allow it to refine these values during bundle adjustment. The output of COLMAP consists of estimated camera intrinsics, extrinsics, and a sparse 3D point cloud.

\subsection{Scale Recovery}

The camera trajectory and point cloud produced by COLMAP are defined only up to an unknown scale factor, an inherent ambiguity of monocular SfM. To resolve this ambiguity, we exploit the metric depth available at each keyframe.

We include the keyframe RGB image in the input to COLMAP so that the keyframe is registered within the same coordinate system as the video frames. Let $\hat{T}_{\text{kf}}$ denote the camera-to-world pose estimated by COLMAP for the keyframe, and let $D$ denote the metric depth map from the structured light sensor. We project the COLMAP sparse point cloud onto the image plane at $\hat{T}_{\text{kf}}$ to obtain an unscaled depth map $\hat{D}$. The scale factor $s$ is then computed as the median of the elementwise ratio between the metric and unscaled depth values:
\begin{equation}
    s = \text{median}\!\left(\frac{D}{\hat{D}}\right).
\end{equation}

Let $\hat{T}_i = (R_i, \hat{t}_i)$ denote the camera-to-world transformation for frame $i$, where $R_i \in \mathrm{SO}(3)$ is the rotation and $\hat{t}_i \in \mathbb{R}^3$ is the camera center in COLMAP coordinates. Because the rotation is scale-invariant, we recover the metric pose by scaling only the translation:
\begin{equation}
    T_i = (R_i,\; s \cdot \hat{t}_i).
\end{equation}

With metric poses in hand, we reproject the associated keyframe depth map into each frame to produce RGB-D pairs. This procedure is repeated independently for every video sequence in the dataset.

\subsection{Limitations and Exclusions}

\textbf{Co-registration requirement.} Only frames that COLMAP successfully co-registers with the keyframe image can be metricized. In some sequences, limited visual overlap or texture leads to low registration rates; for instance, only 11 of the 88 frames in dataset 2, keyframe 1 were co-registered with the keyframe. As a result, the corrected dataset is smaller than the original in terms of total frame count. However, we demonstrate that the corrected frames are more reliable for neural network training. 

\textbf{Datasets 4 and 5.} The original challenge paper~\cite{scared} notes poor calibration for datasets 4 and 5. We attempted to apply the same pipeline to these datasets but were unable to obtain satisfactory results, so we exclude them entirely. After all exclusions, our pipeline produces 17,135 reliable RGB-D pairs. The full breakdown by sequence and suggested train-validation split are provided in Table~\ref{tab:scared_splits} (Appendix).

\section{Experiments}

There is no direct way to evaluate the accuracy of corrected non-keyframe poses, since ground-truth depth exists only at the keyframes. We therefore design two indirect experiments that test whether the corrected data is consistent with the keyframe ground truth. We use 25 of the provided 35 keyframes since datasets 4 and 5 contain imperfect calibration. 
 
\subsection{Stereo Disparity Evaluation}

\textbf{Setup.} We evaluate FoundationStereo~\cite{wen2025foundationstereo}, an off-the-shelf stereo disparity estimation model, on three versions of the dataset: (1) the original SCARED data with kinematics-based poses, (2) our corrected SCARED-C data, and (3) the keyframes only. Metric depth is computed from predicted disparity using the provided stereo calibration via $\text{depth} = (f_x \times \text{baseline}) / \text{disparity}$. We report End-Point Error (EPE) for disparity, along with Absolute Relative error (Abs.\ Rel.) and $\delta_1$ accuracy for depth.

The reasoning behind this experiment is as follows. FoundationStereo predicts disparity from stereo image pairs, independently of the ground-truth depth map. If the corrected dataset contains geometrically accurate RGB-D pairs, then the stereo model's predictions should agree with the reprojected depth maps at a level comparable to its agreement with the keyframe ground truth.

\begin{table}[ht]
\centering
\caption{FoundationStereo evaluation on the original, corrected, and keyframes-only versions of the SCARED data.}
\label{tab:results_1}
\begin{tabular}{lccc}
\toprule
Dataset & EPE $\downarrow$ & Abs.\ Rel.\ $\downarrow$ & $\delta_{1}$ $\uparrow$ \\
\midrule
Original  & 6.062 & 0.046 & 96.3 \\
Corrected & 1.912 & 0.026 & 99.0 \\
\midrule
Keyframes & 0.998 & 0.014 & 99.7 \\
\bottomrule
\end{tabular}
\end{table}

\textbf{Discussion.} Table~\ref{tab:results_1} shows that the corrected dataset substantially closes the gap between the original data and the keyframe ground truth across all three metrics. The EPE drops from 6.062 to 1.912 (a $3.2\times$ reduction), and both Abs.\ Rel.\ and $\delta_1$ approach their keyframe-only values. The remaining gap between the corrected data and the keyframes is expected: reprojected depth maps inevitably contain artifacts from occlusion and interpolation that are absent in the directly measured keyframe depth. We also observed that the original dataset exhibits high variance in per-sequence quality, with some sequences such as dataset 2, keyframe 4 showing catastrophically poor results (EPE of 23.6). The corrected dataset minimizes this error and high variance between sequences. We also report per-keyframe results in Table~\ref{tab:per-keyframe}.


\textbf{Fast-FoundationStereo.} Recently, the authors of FoundationStereo released Fast-FoundationStereo~\cite{wen2025fast}, a faster distilled model for real-time stereo matching. For comprehensive benchmarking, we also compare the two models' performance on the 25 SCARED keyframes, shown in Table~\ref{tab:fastvsfs}. We observe that both models exhibit similar performance with small gains in the original FoundationStereo model. However, Fast-FoundationStereo is over $8\times$ faster. 

\begin{table}[ht]
\centering
\caption{Comparison of FoundationStereo and Fast-FoundationStereo on SCARED keyframes in performance and FPS. The images were processed at original resolution of $1024\times1280$. We used an NVIDIA H200 with batch size $1$ to generate these results.}
\label{tab:fastvsfs}
\begin{tabular}{lcccc}
\toprule
Model & EPE $\downarrow$ & Abs.\ Rel.\ $\downarrow$ & $\delta_{1}$ $\uparrow$ & FPS $\uparrow$\\
\midrule
Fast-FoundationStereo  & 1.011 & 0.014 & 99.6 & \textbf{1.852} \\
FoundationStereo & \textbf{0.998} & \textbf{0.014} &\textbf{99.7}& 0.226\\
\bottomrule
\end{tabular}
\end{table}

\subsection{Monocular Depth Estimation}

\textbf{Setup.} As an alternative method of verification, we train a monocular depth estimation model on either the original or the corrected dataset and evaluate on the 25 keyframes as a held-out test set. Note that keyframe images are not part of any video sequence and are therefore never seen during training. For simplicity, all depth maps are normalized to perform relative depth estimation. We train a ConvNet-based U-Net using the train-validation split specified in Table~\ref{tab:scared_splits} and select the best model based on validation loss. This model is evaluated on the held-out keyframes, whose metrics are reported in Table~\ref{tab:results_2}.

\begin{table}[ht]
\centering
\caption{Relative depth estimation results for models trained on the original and corrected datasets, evaluated on 25 keyframes.}
\label{tab:results_2}
\begin{tabular}{lccc}
\toprule
Dataset & Abs.\ Rel.\ $\downarrow$ & RMSE $\downarrow$ & $\delta_1$ $\uparrow$\\
\midrule
Original  & 0.856 & 0.283 & 18.5 \\
\textbf{Corrected} & \textbf{0.528} & \textbf{0.184} & \textbf{26.3} \\
\bottomrule
\end{tabular}
\end{table}

\textbf{Discussion.} Table~\ref{tab:results_2} confirms that the corrected dataset produces meaningfully better training signal. The model trained on SCARED-C achieves a 38\% lower Abs.\ Rel.\ and a 35\% lower RMSE compared to training on the original data. The $\delta_1$ accuracy improves from 18.5\% to 26.3\%. While both models remain far from the performance ceiling suggested by the stereo evaluation in Table~\ref{tab:results_1} (which achieves near-perfect $\delta_1$), this is expected given the simplicity of the U-Net architecture, the difficulty of monocular depth estimation task, and the dataset size. The key takeaway is that training on the corrected data consistently outperforms training on the original data, indicating that the corrected depth maps provide a more reliable supervisory signal. We emphasize that these results should be interpreted relatively, and not as a SOTA baseline. 

\section{Conclusion}

We have presented SCARED-C, a corrected version of the SCARED endoscopic depth estimation dataset. By replacing the original kinematics-based camera poses with poses estimated through Structure-from-Motion and a simple scale recovery procedure, we expand the number of reliable RGB-D pairs from 35 keyframes to 17,135 frames. Our experiments show that the corrected data is comparable to the keyframe ground truth in stereo evaluation and produces stronger training signal for monocular depth estimation.

The corrected dataset is not without limitations. Frames that COLMAP fails to co-register are lost, and the reprojected depth maps inherit any errors in the sparse reconstruction and scale estimation. Additionally, datasets 4 and 5 remain excluded due to poor calibration. Despite these limitations, we believe that a roughly $490\times$ expansion of reliable labeled data in a real-tissue surgical dataset is a meaningful contribution to the community. Finally, there remains some misalignment between the corrected depth maps and RGB images due to imperfect calibrations and COLMAP performance. 

We release the corrected dataset on HuggingFace along with our scale-recovery code in \url{https://github.com/juseonghan/SCARED-C}, and we encourage the community to adopt the suggested train-validation split in Table~\ref{tab:scared_splits} to enable consistent benchmarking.

\bibliographystyle{unsrt}  
\bibliography{references}  

@article{scared,
  title={Stereo correspondence and reconstruction of endoscopic data challenge},
  author={Allan, Max and Mcleod, Jonathan and Wang, Congcong and Rosenthal, Jean Claude and Hu, Zhenglei and Gard, Niklas and Eisert, Peter and Fu, Ke Xue and Zeffiro, Trevor and Xia, Wenyao and others},
  journal={arXiv preprint arXiv:2101.01133},
  year={2021}
}

@inproceedings{han2026depth,
  title={Depth anything in medical images: A comparative study},
  author={Han, John J and Acar, Ayberk and Henry, Callahan and Wu, Jie Ying},
  booktitle={Medical Imaging 2026: Image-Guided Procedures, Robotic Interventions, and Modeling},
  volume={13927},
  pages={58--66},
  year={2026},
  organization={SPIE}
}

@inproceedings{acar2025monocular,
  title={From monocular vision to autonomous action: Guiding tumor resection via 3d reconstruction},
  author={Acar, Ayberk and Smith, Mariana and Al-Zogbi, Lidia and Watts, Tanner and Li, Fangjie and Li, Hao and Yilmaz, Nural and Scheikl, Paul Maria and d’Almeida, Jesse F and Sharma, Susheela and others},
  booktitle={2025 IEEE/RSJ International Conference on Intelligent Robots and Systems (IROS)},
  pages={21714--21720},
  year={2025},
  organization={IEEE}
}

@article{endoslam,
  title={EndoSLAM dataset and an unsupervised monocular visual odometry and depth estimation approach for endoscopic videos},
  author={Ozyoruk, Kutsev Bengisu and Gokceler, Guliz Irem and Bobrow, Taylor L and Coskun, Gulfize and Incetan, Kagan and Almalioglu, Yasin and Mahmood, Faisal and Curto, Eva and Perdigoto, Luis and Oliveira, Marina and others},
  journal={Medical image analysis},
  volume={71},
  pages={102058},
  year={2021},
  publisher={Elsevier}
}

@article{c3vd,
  title={Colonoscopy 3D video dataset with paired depth from 2D-3D registration},
  author={Bobrow, Taylor L and Golhar, Mayank and Vijayan, Rohan and Akshintala, Venkata S and Garcia, Juan R and Durr, Nicholas J},
  journal={Medical image analysis},
  volume={90},
  pages={102956},
  year={2023},
  publisher={Elsevier}
}

@article{c3vdv2,
  title={C3VDv2--Colonoscopy 3D video dataset with enhanced realism},
  author={Golhar, Mayank V and Fretes, Lucas Sebastian Galeano and Ayers, Loren and Akshintala, Venkata S and Bobrow, Taylor L and Durr, Nicholas J},
  journal={arXiv preprint arXiv:2506.24074},
  year={2025}
}

@article{realsyncol,
  title={RealSynCol: a high-fidelity synthetic colon dataset for 3D reconstruction applications},
  author={Lena, Chiara and Milesi, Davide and Casella, Alessandro and Carlini, Luca and Norton, Joseph C and Martin, James and Scaglioni, Bruno and Obstein, Keith L and De Sire, Roberto and Spadaccini, Marco and others},
  journal={arXiv preprint arXiv:2602.08397},
  year={2026}
}

@inproceedings{martyniak2025simuscope,
  title={Simuscope: Realistic endoscopic synthetic dataset generation through surgical simulation and diffusion models},
  author={Martyniak, Sabina and Kaleta, Joanna and Dall'Alba, Diego and Naskr{\k{e}}t, Micha{\l} and P{\l}otka, Szymon and Korzeniowski, Przemys{\l}aw},
  booktitle={2025 IEEE/CVF Winter Conference on Applications of Computer Vision (WACV)},
  pages={4268--4278},
  year={2025},
  organization={IEEE}
}

@inproceedings{wen2025foundationstereo,
  title={Foundationstereo: Zero-shot stereo matching},
  author={Wen, Bowen and Trepte, Matthew and Aribido, Joseph and Kautz, Jan and Gallo, Orazio and Birchfield, Stan},
  booktitle={Proceedings of the IEEE/CVF conference on computer vision and pattern recognition},
  pages={5249--5260},
  year={2025}
}

@article{wen2025fast,
  title={Fast-FoundationStereo: Real-Time Zero-Shot Stereo Matching},
  author={Wen, Bowen and Dewan, Shaurya and Birchfield, Stan},
  journal={arXiv preprint arXiv:2512.11130},
  year={2025}
}

@inproceedings{colmap,
    author={Sch\"{o}nberger, Johannes Lutz and Frahm, Jan-Michael},
    title={Structure-from-Motion Revisited},
    booktitle={Conference on Computer Vision and Pattern Recognition (CVPR)},
    year={2016},
}

@inproceedings{han2025endopbr,
  title={EndoPBR: Photorealistic Synthetic Data for Surgical 3D Vision via Physically-based Rendering},
  author={Han, John J and Wu, Jie Ying},
  booktitle={Proceedings of the IEEE/CVF Winter Conference on Applications of Computer Vision},
  pages={5601--5611},
  year={2026}
}

\clearpage
\appendix
\begin{table}
\centering
\caption{Suggested train and validation split for the corrected SCARED dataset. The dataset is split approximately 70-30. The sequence code follows the format \{dataset\}\_\{keyframe\}.}
\label{tab:scared_splits}
\begin{tabular}{lrlr}
\toprule
\multicolumn{2}{c}{\textbf{Train}} & \multicolumn{2}{c}{\textbf{Validation}} \\
\cmidrule(lr){1-2} \cmidrule(lr){3-4}
\textbf{Seq} & \textbf{Frames} & \textbf{Seq} & \textbf{Frames} \\
\midrule
1\_1 & 197   & 1\_2 & 280 \\
1\_3 & 471   & 1\_4 & 1 \\
2\_2 & 1,033 & 1\_5 & 1 \\
2\_4 & 2,114 & 2\_1 & 11 \\
3\_1 & 329   & 2\_3 & 1,102 \\
3\_2 & 1,597 & 2\_5 & 1 \\
3\_3 & 448   & 3\_4 & 834 \\
6\_1 & 637   & 3\_5 & 1 \\
6\_2 & 1,087 & 6\_4 & 1,360 \\
6\_3 & 1,573 & 6\_5 & 1 \\
7\_1 & 647   & 7\_2 & 628 \\
7\_4 & 2,197 & 7\_3 & 584 \\
     &       & 7\_5 & 1 \\
\midrule
\textbf{Total} & \textbf{12,330} & \textbf{Total} & \textbf{4,805} \\
\bottomrule
\multicolumn{4}{c}{\textbf{Grand Total: 17,135}} \\
\end{tabular}
\end{table}

\begin{table}[htbp]
\centering
\caption{Per-keyframe evaluation on the corrected SCARED dataset with FoundationStereo. Predicted disparity is converted to depth using the provided stereo calibration. Disparity metrics are in pixels; depth RMSE and MAE are in mm.}
\label{tab:per-keyframe}
\begin{adjustbox}{max width=\textwidth}
\begin{tabular}{
  c
  S[round-precision=2] S[round-precision=2] S[round-precision=1]
  S[round-precision=3] S[round-precision=2] S[round-precision=2] S[round-precision=2]
}
\toprule
 & \multicolumn{3}{c}{\textbf{Disparity}} & \multicolumn{4}{c}{\textbf{Depth}} \\
\cmidrule(lr){2-4} \cmidrule(lr){5-8}
{Dataset / Keyframe}
  & {EPE\,$\downarrow$} & {RMSE\,$\downarrow$} & {Bad\,3\,$\downarrow$}
  & {Abs\,Rel\,$\downarrow$} & {RMSE\,$\downarrow$} & {MAE\,$\downarrow$} & {$\delta<1.25$\,$\uparrow$} \\
\midrule
1 / 1 & 0.75 & 1.00 &  0.2 & 0.011 & 1.04 & 0.67 & 99.99 \\
1 / 2 & 0.67 & 1.46 &  0.3 & 0.012 & 1.44 & 0.87 & 99.89 \\
1 / 3 & 0.62 & 1.16 &  0.2 & 0.011 & 1.26 & 0.85 & 100.00 \\
1 / 4 & 1.19 & 1.68 &  0.9 & 0.012 & 0.75 & 0.53 & 99.99 \\
1 / 5 & 0.83 & 1.27 &  0.5 & 0.010 & 1.05 & 0.59 & 99.99 \\
\midrule
2 / 1 & 1.62 & 2.47 &  1.2 & 0.017 & 1.21 & 0.76 & 99.98 \\
2 / 2 & 0.85 & 1.36 &  0.8 & 0.017 & 2.00 & 1.41 & 99.97 \\
2 / 3 & 1.97 & 3.12 &  6.7 & 0.021 & 1.89 & 1.07 & 99.75 \\
2 / 4 & 1.08 & 1.85 &  0.2 & 0.010 & 0.76 & 0.37 & 99.97 \\
2 / 5 & 1.02 & 2.28 &  1.4 & 0.017 & 2.66 & 1.30 & 99.55 \\
\midrule
3 / 1 & 0.91 & 1.68 &  3.4 & 0.015 & 2.54 & 1.31 & 99.85 \\
3 / 2 & 1.05 & 2.00 &  1.2 & 0.014 & 3.01 & 0.98 & 99.57 \\
3 / 3 & 0.81 & 1.40 &  1.1 & 0.014 & 2.54 & 1.14 & 99.73 \\
3 / 4 & 0.84 & 1.60 &  1.0 & 0.012 & 2.78 & 0.84 & 99.71 \\
3 / 5 & 2.67 & 10.79 & 10.7 & 0.035 & 5.71 & 2.20 & 98.51 \\
\midrule
6 / 1 & 0.55 & 1.64 &  0.2 & 0.012 & 2.10 & 1.26 & 99.97 \\
6 / 2 & 0.78 & 1.78 &  0.3 & 0.011 & 1.52 & 0.72 & 99.97 \\
6 / 3 & 0.67 & 1.14 &  0.7 & 0.011 & 1.47 & 0.86 & 99.99 \\
6 / 4 & 0.77 & 2.48 &  1.8 & 0.013 & 3.04 & 1.19 & 98.89 \\
6 / 5 & 1.06 & 4.50 &  2.1 & 0.018 & 5.56 & 1.74 & 98.51 \\
\midrule
7 / 1 & 1.05 & 1.80 &  1.2 & 0.016 & 2.44 & 1.38 & 99.90 \\
7 / 2 & 1.13 & 2.39 &  0.9 & 0.014 & 1.81 & 0.97 & 99.87 \\
7 / 3 & 1.46 & 5.42 &  1.2 & 0.012 & 1.43 & 0.53 & 99.63 \\
7 / 4 & 1.10 & 2.08 &  2.8 & 0.013 & 2.16 & 0.80 & 99.78 \\
7 / 5 & 0.65 & 3.67 &  0.6 & 0.011 & 2.32 & 0.86 & 99.91 \\
\midrule
\textbf{Mean} & \textbf{1.00} & \textbf{3.05} & \textbf{1.6}
              & \textbf{0.014} & \textbf{2.54} & \textbf{1.02} & \textbf{99.71} \\
\bottomrule
\end{tabular}
\end{adjustbox}
\end{table}

\end{document}